\documentclass{llncs}
\usepackage{graphicx}
\usepackage{amsmath,amssymb}
\usepackage{bm}
\usepackage{color}
\usepackage[width=135mm,left=23mm,paperwidth=176mm,height=210mm,top=18mm,paperheight=246mm]{geometry}

\usepackage[backend=biber, sorting=none]{biblatex}
\usepackage{hyperref}
\usepackage[parfill]{parskip}

\begin{document}

\pagestyle{headings}

\mainmatter

\title{Supervised Transfer Learning Framework for Fault Diagnosis in Wind Turbines}

\author{Kenan Weber \and Christine Preisach}
\institute{Hochschule Karlsruhe - University of Applied Sciences\\ 
\email{kenan.weber@h-ka.de}\\
\email{christine.preisach@h-ka.de}
}

\maketitle

\begin{abstract}

Common challenges in fault diagnosis include the lack of labeled data and the need to build models for each domain, resulting in many models that require supervision.
Transfer learning can help tackle these challenges by learning cross-domain knowledge. Many approaches still require at least some labeled data in the target domain, and often provide unexplainable results. 
To this end, we propose a supervised transfer learning framework for fault diagnosis in wind turbines that operates in an \textit{Anomaly-Space}. This space was created using SCADA data and vibration data and was built and provided to us by our research partner. Data within the Anomaly-Space can be interpreted as anomaly scores for each component in the wind turbine, making each value intuitive to understand.
We conducted cross-domain evaluation on the train set using popular supervised classifiers like Random Forest, Light-Gradient-Boosting-Machines and Multilayer Perceptron as metamodels for the diagnosis of bearing and sensor faults. The Multilayer Perceptron achieved the highest classification performance. This model was then used for a final evaluation in our test set. The results show, that the proposed framework is able to detect cross-domain faults in the test set with a high degree of accuracy by using one single classifier, which is a significant asset to the diagnostic team.

\keywords{Condition Monitoring, Wind Turbines, Anomaly Detection, Fault Detection, Fault Diagnosis, Transfer Learning}
\end{abstract}

\section{Introduction}
In Germany, electricity generated from wind turbines (WTs) makes up a large portion of the total generated energy from renewable energy sources \cite{de_windenergie}.
In order to increase the total energy yield, it is very important to reduce total downtimes by monitoring critical aspects of WTs. With condition monitoring, faults can be detected early and maintenance times and measures can be planned accordingly. This further reduces the risk of total failure due to the propagation of the faults to other areas of the machine.
The total amount of WTs is increasing, with each of them equipped with more and more sensors over the years. Hence, this raises the amount of components, that can be measured which leads to higher costs and more signals, that need to be monitored. Furthermore, when a fault has been detected, we still need to infer the type of fault and localize it, e.g. which component(s) is/are affected. With an increasing amount of signals, there is need for more and more highly specialized personnel in order to monitor these machines manually. Therefore, we propose a solution that automates this whole process from end-to-end.

Supervised learning is a method for performing intelligent fault diagnosis on WTs. In literature, numerous solutions based on supervised learning are available \cite{multiscale_cnn,gpc,elm}. However, these solutions typically focus solely on diagnosing faults in one particular machine, resulting in the development of  separate models for each machine. Transfer learning is a promising approach to use knowledge extracted from a subset of WTs to multiple other ones. This can decrease the total model count needed for reliable fault diagnosis. Additionally, transfer learning based fault diagnosis is not limited to detect only those fault types, that have been occurred in the past on that particular WT, since fault data from other WTs can also be used for the diagnosis. Many transfer learning solutions for WTs exist. Zhang et al. \cite{zhang2018wind} built a fully connected neural network which is able to detect, whether ice are on the WT blades by only using SCADA data. A small data set from another WT was used to fine-tune the model. Yang et al. \cite{yang2021image} are able to detect blade defects by segmenting blade images with the otsu threshold segmentation algorithm and then using a pre-trained Alexnet classifier for the feature extraction. A Random Forest was used for the fault diagnosis in the last step.
Li et al. \cite{li2021wind} pre-trained
a convolutional autoencoder on SCADA data from 14 WTs and fine-tuned the model on data from the 15th WT in order to detect fault types like high temperature in gearbox or generator or low pressure of the hydraulic system. A stacked autoencoder was employed by Deng et al. \cite{deng2021deep} by pretraining the model on source data and then utilizing and fine-tuning a fully connected layer for the diagnosis in the target domain. 

The aforementioned solutions have the limitation, that some labeled data in the source domain have to be available. Furthermore, features extracted with the help of neural networks are mostly abstract and not interpretable for diagnosticians. If the fault diagnosis system indicates that a fault is present, then it has to be clear to the diagnostician, \textit{how} the decision was made.
To this end, we propose a fault diagnosis framework, which operates in an \textit{Anomaly-Space}. This new feature space provides several normalized anomaly scores for each WT component, for every available WT. Values above 1.0 are considered anomalous. It was built and provided to us by our research partner EnBW Energie Baden-Württemberg AG. Both SCADA data and vibration data were used for the creation of the Anomaly-Space. A supervised classifier takes data from the Anomaly-Space as input and provides fault diagnosis results. These results can easily be interpreted by diagnosticians, since features in the Anomaly-Space represent deviations from the normal behavior of the WT. This can be seen as a feature-based transfer learning approach, where the Anomaly-Space represents the domain-shared feature space.

In summary, the contributions of this paper are the following:

\begin{enumerate}
    \item Fault diagnosis based on  derived signals from SCADA data and vibration data, that are easily interpretable, in contrast to many other transfer learning approaches.
    \item Extensive model training and evaluation with stratified cross-validation from real data across 5 WTs from 4 wind parks and comparing classification performance of popular supervised classifiers, such as Random Forest (RF), Light-Gradient-Boosting-Machines (LightGBM) and Multilayer Perceptron (MLP).
    \item Showing transfer learning capabilities by applying  the best performing classifier from the aforementioned analysis on a new test set, which consists of 2 WTs from 2 wind parks, one of which is a completely different wind park compared to the train set.
\end{enumerate}

This paper is organized as follows: In \autoref{sec:data_2}, we give an overview about the dataset. This includes a brief description and explanation of the Anomaly-Space. In \autoref{sec:fd_general_3}, some background information about transfer learning in fault diagnosis is given. Our supervised fault diagnosis framework is introduced in \autoref{sec:supervised_fd_4}. Results are shown in \autoref{sec:results_5} and 
conclusions are made in \autoref{sec:concl_6}.

\section{Dataset Description}
\label{sec:data_2}
  The dataset we used contains fault-types and anomaly scores, which are deduced from SCADA data and vibration data. Two common faults can be found within the data: bearing fault and sensor fault.
Bearing faults are common and severe faults in WTs. Ignoring these can result in a total failure of the machine and leads to substantial down-times and repair costs. Therefore, there is a huge interest in detection of bearing faults as early as possible. 

Faulty sensors are also very common in WTs, but are very cheap to replace in terms of raw material cost and are not damaging the WT itself. At a first glance, this might not be as important, but these type of faults can lead to several problems. 
Unnecessary maintenance works could be performed if sensor faults are confused with a more serious disturbance. Consequently, diagnosticians and technicians might increasingly distrust the fault diagnosis application and/or mistake a serious fault with a sensor fault.

SCADA data are typically used for condition monitoring in WTs. In general, the term SCADA stands for "Supervisory Control and Data Acquisition" and refers to the monitoring and control of technical processes using data that originates from sensors, actuators and other field devices and is sent to a control system. Among other things, process variables such as temperature, pressure and similar values are recorded. Each recorded 10-minute window is aggregated into four scalar values: minimum, maximum, standard deviation and average. Vibration sensors are able to capture data in a much higher sample rate than SCADA and are commonly used to identify early signs of wear, imbalance, or misalignment in rotating machinery.

The data originates from a total of 7 WTs, across 5 different wind parks. All fault cases with further information are listed in \autoref{table:dataset}.
Our train set (case 1 to 6) consists of data from 5 WTs, which are from 4 different wind parks.
The test set (case 7 and 8) has data from 2 WTs, one of which is from a completely different wind park.

\setlength{\tabcolsep}{4pt}
\begin{table}
\begin{center}
\caption{More information about the data. The train data is considered the source data, validation and test data are considered target data.  P = park, U = unit, (N)DE = (Non-)Drive End.}
\label{table:dataset}
\begin{tabular}{cccccc}
\hline\noalign{\smallskip}
Park/Unit & Fault-Type & Fault-Location & Dataset & Case-No.\\
\noalign{\smallskip}
\hline
\noalign{\smallskip}
P1/U1 & sensor fault & temperature generator phase 3 & train/validation & 1\\
P2/U1 & sensor fault & temperature transformator phase 3 & train/validation & 2\\
P2/U1 & sensor fault & temperature generator phase 1 & train/validation & 3\\
P3/U1 & bearing fault & Fast Shaft Bearing DE & train/validation & 4\\
P3/U2 & bearing fault & Fast Shaft Bearing NDE & train/validation & 5\\
P4/U1 & bearing fault & Fast Shaft Bearing DE & train/validation & 6\\
\vspace{1pt} \\
P1/U2 & sensor fault & temperature generator phase 2 & test & 7\\
P5/U1 & bearing fault & Fast Shaft Bearing DE & test & 8\\

\hline
\end{tabular}
\end{center}
\end{table}
\setlength{\tabcolsep}{1.4pt}

\subsection{Anomaly-Space}

The Anomaly-Space refers to a feature space, that has been constructed by using multiple proprietary algorithms, denoted as detectors. The input of these detectors are SCADA data and vibration data.

Detectors are monitoring critical components of the WT and provide anomaly scores, which represent a deviation from the normal behavior. These values are normalized such that measurements with values above 1.0 are considered anomalous.

Broad-Band-Characteristic-Value (bbcv) is one of the detectors. This detector first captures multiple windows of vibration data when pre-defined conditions have been met (e.g. approximately constant wind speeds). Several features are then extracted from the raw vibration data and from the frequency domain, after applying the Fast-Fourier-Transformation (FFT), such as skewness, kurtosis and average values.
In the last step, the trendiness of the aforementioned features with the help of hypothesis testing is being calculated.
Bearing faults will usually result in an increased trendiness in multiple features.

Another detector is the tuplet detector. This detector is designed to detect SCADA data deviations from groups of semantically similar components, for example the generator temperature of all three voltage supply phases. This is being achieved by monitoring the variance of the measurements of these components. A sensor fault in any of these components significantly increases the variance. A statistical test quantifies the difference to the expected null hypothesis, null hypothesis being a variance value of 0.

\begin{figure}[ht!]
    \centering
    \begin{minipage}{0.45\textwidth}
        \centering
        \includegraphics[width=\textwidth]{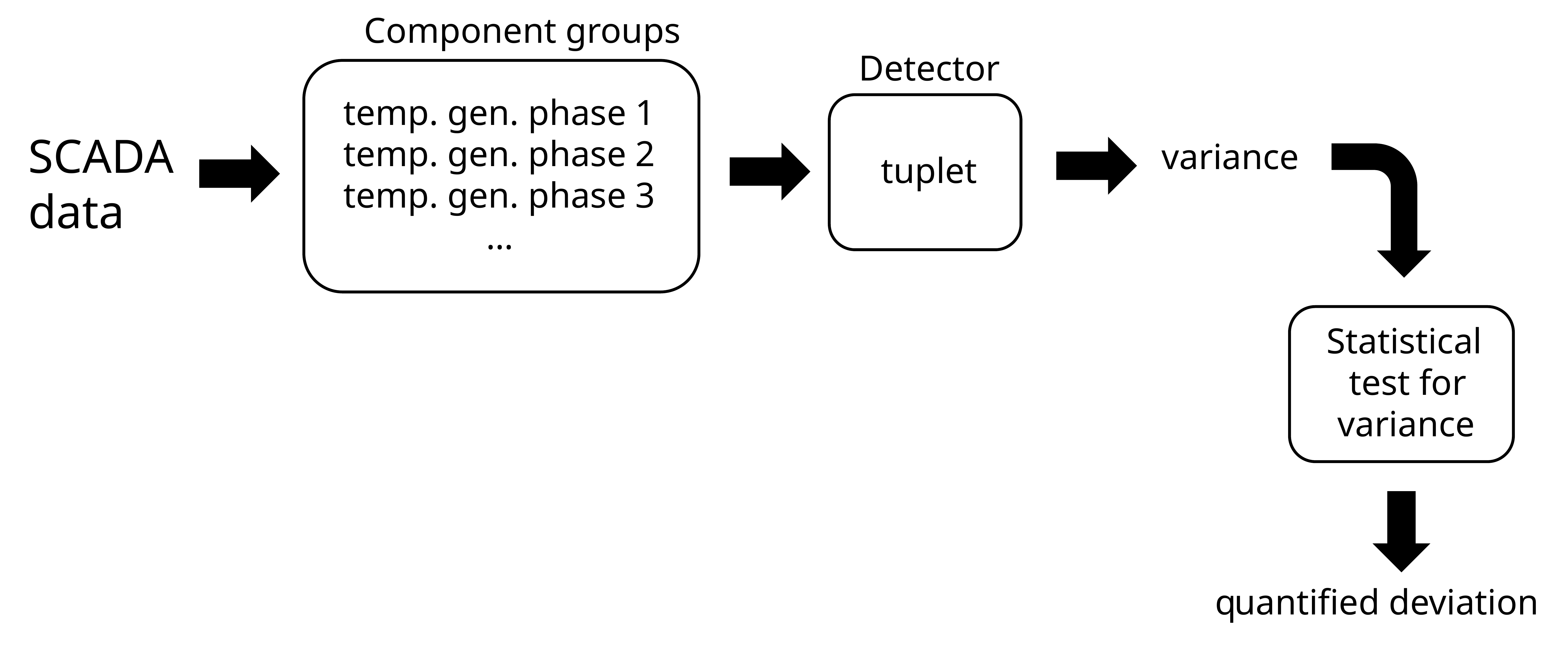}
        \caption{tuplet detector procedure}
        \label{fig:tuplet}
    \end{minipage}
    \hfill
    \begin{minipage}{0.45\textwidth}
        \centering
        \includegraphics[width=\textwidth]{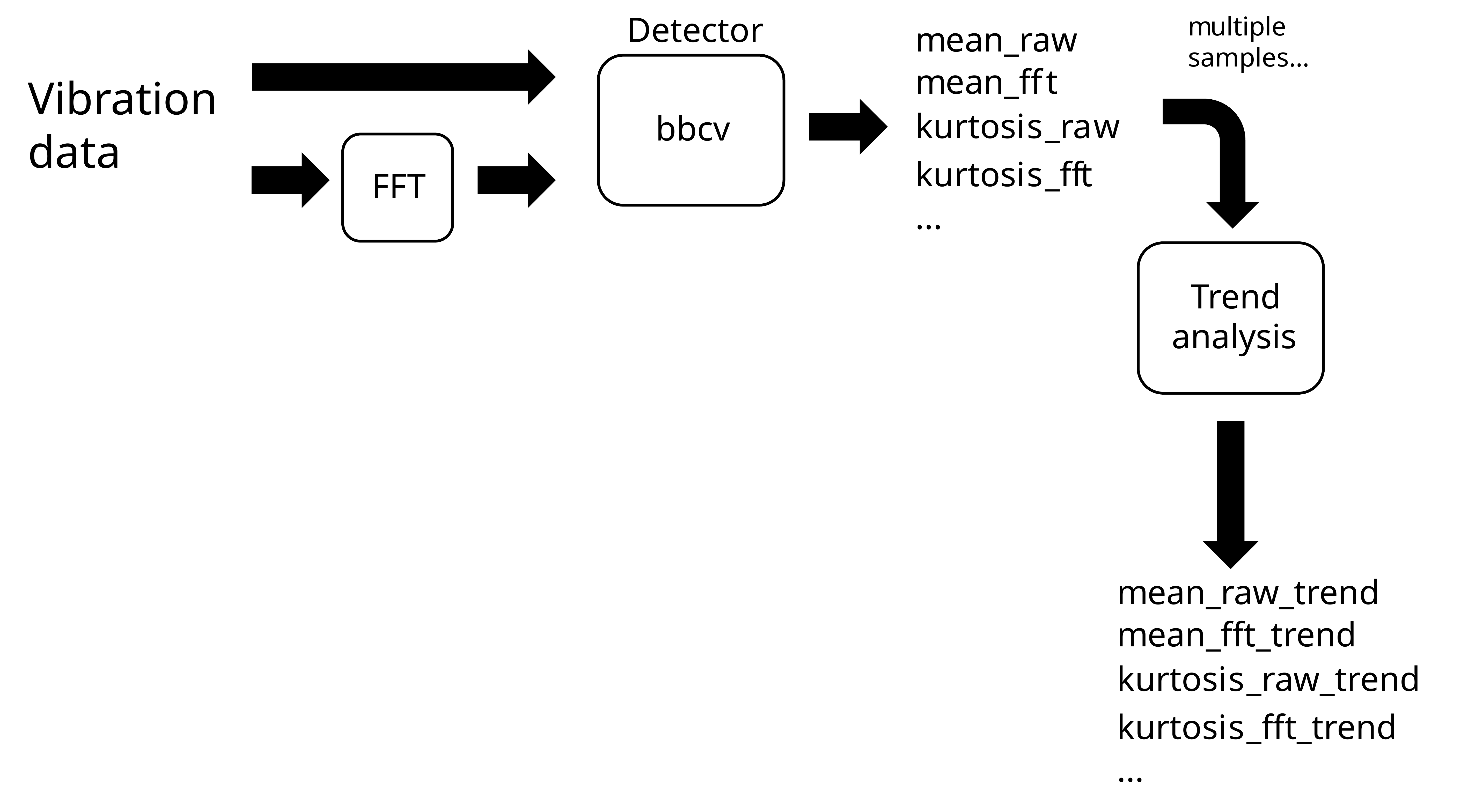}
        \caption{bbcv detector procedure}
        \label{fig:bbcv}
    \end{minipage}
\end{figure}

The procedure of both detectors are depicted in \autoref{fig:tuplet} and \autoref{fig:bbcv}.

The Anomaly-Space is provided to us by our research partner EnBW Energie Baden-Württemberg AG and is used as input data to our preprocessing steps and metamodels.

\section{Transfer Learning in Fault Diagnosis}
\label{sec:fd_general_3}

Fault diagnosis can be decomposed into fault detection and fault classification. 
Fault detection (e.g. using anomaly/outlier detection) is the detection of deviations from normal behavior in the data. These deviations come from either a fault within the monitored system, or from a faulty measuring device. The transformation of the input data can help in detecting faults, that are otherwise not visible. 

To be able to diagnose the type of fault, the transformed data can then be classified into pre-defined fault-classes (labels) using supervised classification algorithms.

One of the challenges of fault diagnosis is the lack of labels, since faults occur rarely. In the context of WTs, there might be many fault-labels available across all available WTs. However, there are many reasons why it is still not eligible to train a ML model on one WT and apply it on another without some modifications. For example, WTs may come from different manufacturers and therefore have different components, which results in non-similar signal patterns. Consequently, at least one model for each WT needs to be created.

Transfer learning aims to capture domain-invariant knowledge from just few available domains and apply this knowledge onto other domains. In the context of WTs, for instance, the objective is to leverage insights gained from a few WTs and extend them to others.

 There are 4 types of transfer learning methods according to Lei et al. \cite{lei2020applications}: \textbf{feature-based} approaches, \textbf{GAN-based} approaches, \textbf{instance-based} approaches and \textbf{parameter-based} approaches. Feature-based approaches map cross-domain data into a common feature space and decrease the distribution discrepancy before applying a classifier. GAN-based approaches utilize the GAN framework in order to learn the distribution of the target data and generate new ones to improve the classifier. Instance-based approaches reweight misclassified instances from source and target domain, increasing/decreasing the influence of those instances on the fault diagnosis classifier. Parameter-based approaches train models (e.g. neural networks) on the source data and fine-tune the learned model parameters on the target data. Instance- and parameter-based approaches assume, that few labeled samples of the target domain are available.
 Our solution can be regarded as a feature-based TL approach without the step of further decreasing the distribution discrepancy. This space is designed to represent deviations from the individual normal behavior, making each value intuitive to understand.

\section{Supervised Fault Diagnosis Framework}
\label{sec:supervised_fd_4}

\begin{figure}[ht!]
    \centering
    \includegraphics[width=0.7\textwidth]{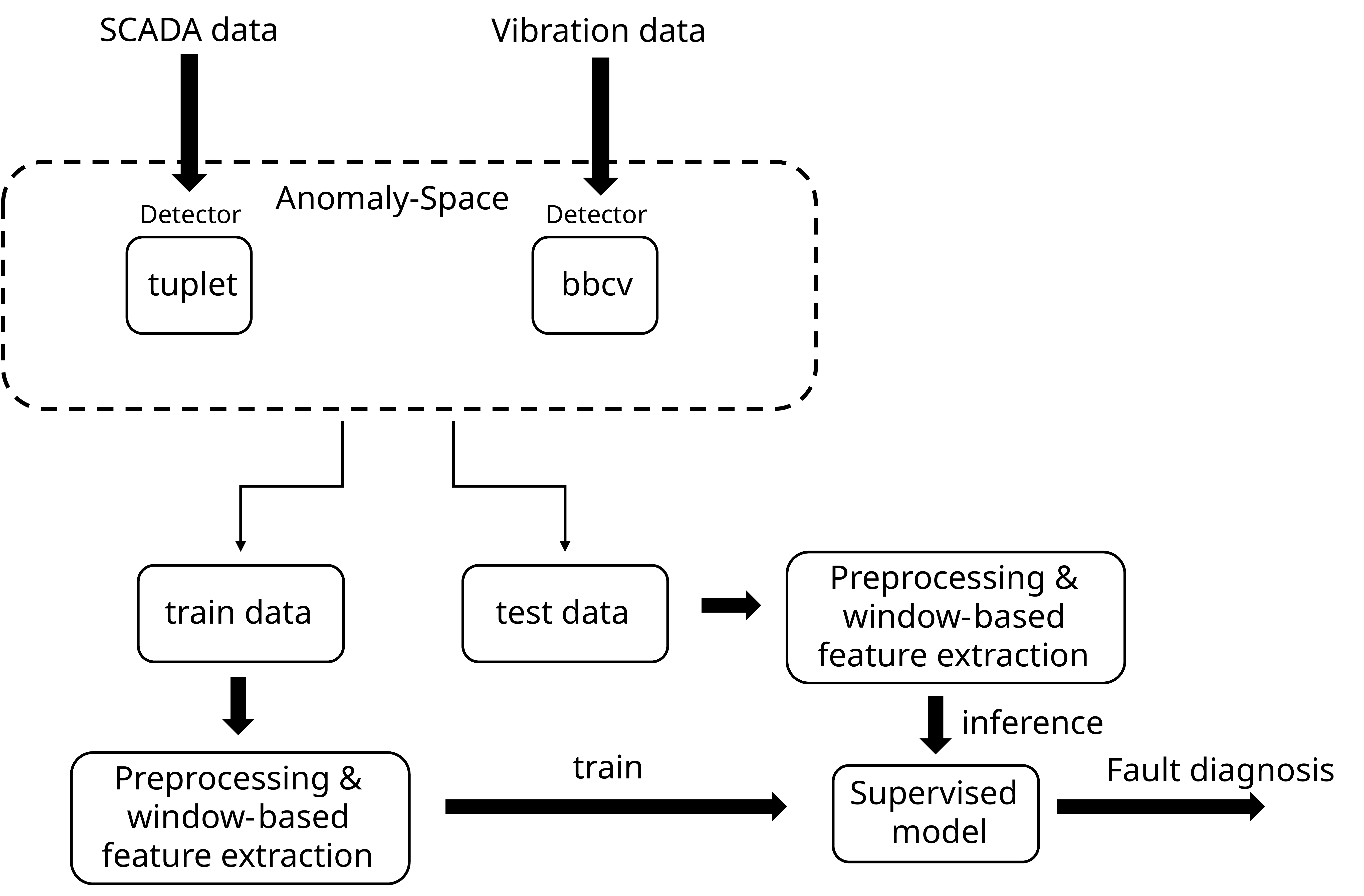}
    \caption{Proposed fault diagnosis framework}
    \label{fig:fd_framework}
\end{figure}

A general overview of the proposed fault diagnosis framework is depicted in \autoref{fig:fd_framework}.

The dataset labels are generated using fault time frames (appearance dates and repair dates) provided by diagnosticians. Data from within the fault time frames are labeled with the corresponding fault type. Data outside these time frames are labeled as "Normal". Data that are not within the normal operating-mode of the WT (e.g. stillstand or wind speeds below a specified threshold) have been omitted. A time-based forward fill was applied, filling missing values for up to 3 hours after the first occurrence. Remaining data gaps are filled with the value 0.0.

The resulting data frame has data from each available WT component as observations (rows) with each detector output being a feature (column). 

The amount of features provided by the bbcv detector is reduced to one single feature, by only keeping the feature with the largest variance. Consequently, both detectors provide only a single feature each, resulting so far in 2 total features.

Sliding-window based feature extraction has been employed in order to capture the relationship between neighbored data samples, with a window-size of 144 and a stride of 1. The extracted features are trend-certainty (tc) and variance (var). The Mann-Kendall test is a statistical test for determining trends in data. 
We used \autoref{eq:1} to determine the value of our trend-certainty feature, with $p_{mk}$ being the p-value for having a positive trend within a window. The values of both features have been manually set to 0.0, if the window contains only values below 1.0. The final feature count has been increased to 6 after extracting those features.

For MLP, the base features, e.g. the outputs of both detectors, have been normalized with a min-max scaler.

\begin{align}
\label{eq:1}
tc =
    \begin{cases} 
  1 & \text{if } p_{mk} < 0.001, \\
    0 & \text{otherwise.}
    \end{cases}
\end{align}
\section{Results \& Discussion}
\label{sec:results_5}
Stratified 3-fold cross-validation was used for evaluation on the train data. We chose the $f_{\beta}$-score with $\beta = 0.5$ as our evaluation metric. This way, precision has a larger impact than recall. This choice was made because trusting the fault diagnosis is crucial, which can be achieved by minimizing false positive predictions; therefore, precision should be weighted more heavily. Additionally, measurements from faulty sensors could occasionally resemble healthy ones, if the root-cause is loose contact. There could be multiple days of data within the fault time frame, where the sensor delivers non-faulty measurements. Weighting precision and recall equally (e.g. with $\beta = 1$) would provide overly pessimistic evaluation values in these cases.

\subsection{Comparison between different classifiers on train data}

\begin{figure}[ht!]
    \centering
    \includegraphics[width=0.7\textwidth]{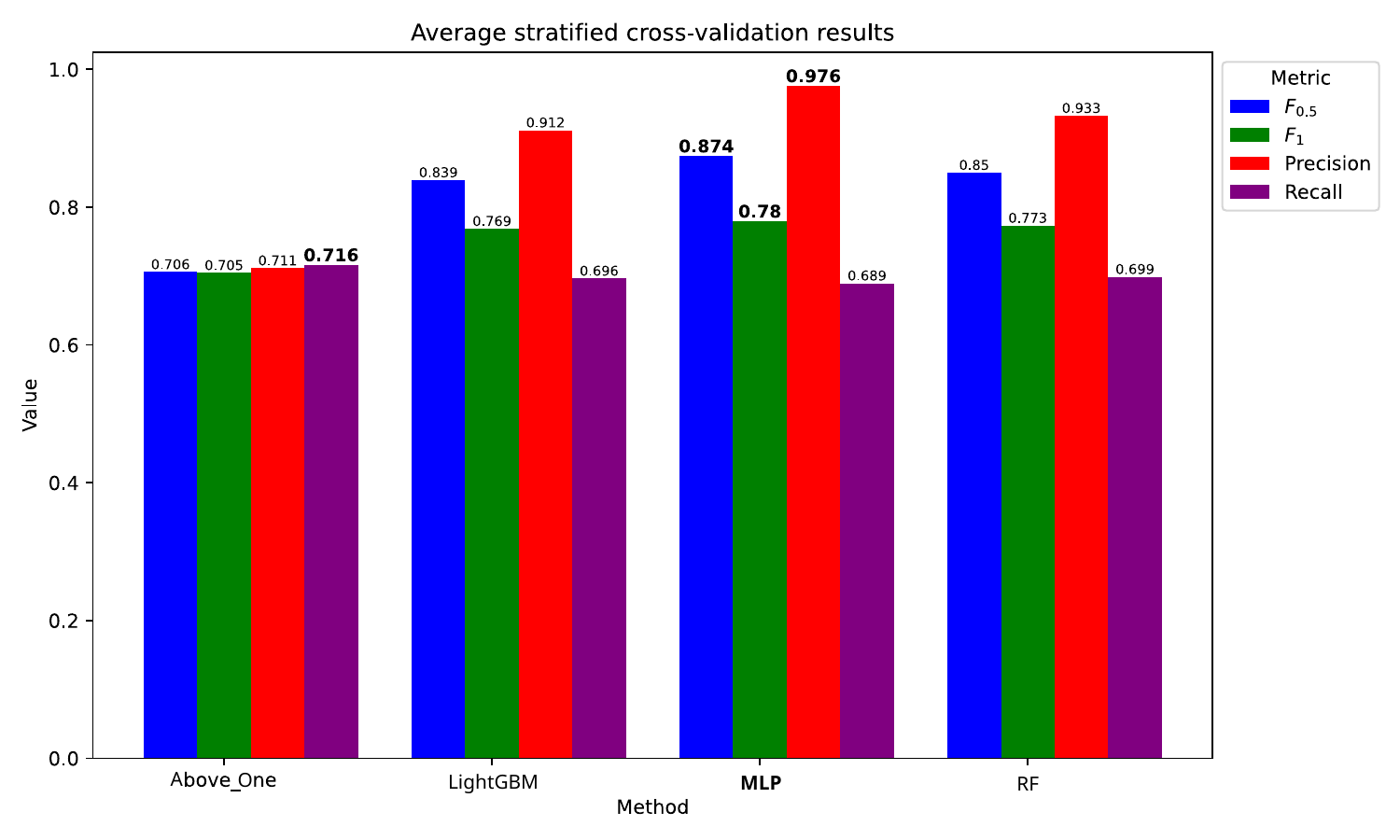}
    \caption{Average stratified cross-validation results}
    \label{fig:cv_results}
\end{figure}

Several popular classifiers were used for our evaluation on the train data: RF, LightGBM and MLP. We created a baseline model which classifies each instance with bbcv values above 1.0 as bearing fault and tuplet values above 1.0 as sensor fault, which we simply termed \textit{Above-One}.
The results are depicted in \autoref{fig:cv_results}. The best performing model is MLP, achieving a $\bm{F_{0.5}}$ score of \textbf{0.874} with the following hyperparameters: ReLU activation function, Adam optimizer, learning rate of 0.001 and 1 hidden layer with 5 neurons.

\subsection{Evaluating the best classifier on test data}

The MLP was trained on the whole train data with the best parameters and then applied on the test data. \autoref{table:test_results} shows the results. The model achieved a $\bm{F_{0.5}}$ score of \textbf{0.937}.

\setlength{\tabcolsep}{4pt}
\begin{table}
\begin{center}
\caption{Evaluation results on test data.}
\label{table:test_results}
\begin{tabular}{llllll}
\hline\noalign{\smallskip}
Method & $F_{0.5}$ score & $F_{1}$ score & Precision & Recall & \\
\noalign{\smallskip}
\hline
\noalign{\smallskip}
MLP & \textbf{0.937} & 0.871 & 0.992 & 0.789 & \\
\hline
\end{tabular}
\end{center}
\end{table}
\setlength{\tabcolsep}{1.4pt}

\subsection{Discussion}
It can be seen from \autoref{fig:cv_results} and \autoref{table:test_results}, that higher evaluation scores have been achieved in the test data (0.874 vs. 0.937). This is due to data quality issues, that are present in the train data. More precisely, there is a loose contact sensor fault case, in which multiple signal segments appear normal, within the fault time frame. These segments can be present multiple days. This complicates the fault diagnosis evaluation, since our window size for our feature extraction methods is approximately one day. Possible solutions would be to increase the window size or to split the fault time frame into multiple smaller ones to increase the coverage of the visible fault pattern.

\section{Conclusion \& Future Work}
\label{sec:concl_6}
In this paper, a fault diagnosis framework based on an Anomaly-Space is proposed. The Anomaly-Space is a feature space, in which deviations from normal behavior (anomaly scores) for each WT component are encoded. Window-based features are then extracted from the Anomaly-Space, such as trend values extracted with the Mann-Kendall test. This new feature space provides intuitive values which can help explain fault diagnosis results, since these values represent deviations from the normal behavior in contrast to many other approaches.
This framework can be regarded as a feature-based transfer learning method without further decreasing the distribution discrepancy.
Supervised classifiers such as Random Forest, Light-Gradient-Boosting-Machines and Multilayer Perceptron are compared on the train data with stratified cross-validation. The Multilayer Perceptron achieved the highest classification performance in diagnosing bearing and sensor faults and was tested on 2 new WTs, one of which stems from a different wind park, compared to the train data. This final evaluation also showed good results, making this a promising fault diagnosis approach for cross-domain fault diagnosis.
Future work could include Out-Of-Distribution (OOD) detection to the framework, in order to detect previously unseen fault types.

\section*{Acknowledgement}

This work was conducted as part of the research project  \textit{AutoDiagCM - Automatisierte Diagnose von Schäden an Windenergieanlagen} (grantnumber 03EE2046B) funded by the German Federal Ministry of Economic Affairs and Climate Action and in cooperation with our research partner EnBW Energie Baden-Württemberg AG, who kindly provided us with data from the Anomaly-Space.

\printbibliography

@misc{de_windenergie,
  title = {Erneuerbare Energien in Deutschland Daten zur Entwicklung im Jahr 2023},
  author = {Geschäftsstelle der Arbeitsgruppe Erneuerbare Energien-Statistik (AGEE-Stat) am Umweltbundesamt},
  url = {https://www.umweltbundesamt.de/publikationen/erneuerbare-energien-in-deutschland-2023},
  note = {Accessed: 7-19-2024},
}

@article{gpc,
  title = {Wind turbine fault diagnosis based on Gaussian process classifiers applied to operational data},
  author = {Yanting Li and Shujun Liu and Lianjie Shu},
  journal = {Renewable Energy},
  volume = {134},
  pages = {357-366},
  year = {2019},
  issn = {0960-1481},
  doi = {10.1016/j.renene.2018.10.088}
}

@article{multiscale_cnn,
  title={Multiscale Convolutional Neural Networks for Fault Diagnosis of Wind Turbine Gearbox}, 
  author={Jiang, Guoqian and He, Haibo and Yan, Jun and Xie, Ping},
  journal={IEEE Transactions on Industrial Electronics}, 
  volume={66},
  number={4},
  pages={3196-3207},
  year={2019},
  doi={10.1109/TIE.2018.2844805}
}

@article{elm,
  title = {Representational Learning for Fault Diagnosis of Wind Turbine Equipment: A Multi-Layered Extreme Learning Machines Approach},
  author = {Yang, Zhi-Xin and Wang, Xian-Bo and Zhong, Jian-Hua},
  journal = {Energies},
  volume = {9},
  number = {6},
  article-number = {379},
  year = {2016},
  issn = {1996-1073},
  doi = {10.3390/en9060379}
}

@inproceedings{zhang2018wind,
  title = {Wind turbine ice assessment through inductive transfer learning},
  author = {Zhang, Chengkai and Bin, Junchi and Liu, Zheng},
  booktitle = {2018 ieee international instrumentation and measurement technology conference (i2mtc)},
  pages = {1--6},
  year = {2018},
  organization = {IEEE},
  doi = {10.1109/I2MTC.2018.8409794}
}

@article{yang2021image,
  title={Image recognition of wind turbine blade damage based on a deep learning model with transfer learning and an ensemble learning classifier},
  author={Yang, Xiyun and Zhang, Yanfeng and Lv, Wei and Wang, Dong},
  journal={Renewable Energy},
  volume={163},
  pages={386--397},
  year={2021},
  publisher={Elsevier},
  doi={10.1016/j.renene.2020.08.125}
}

@article{li2021wind,
  title={Wind turbine fault diagnosis based on transfer learning and convolutional autoencoder with small-scale data},
  author={Li, Yanting and Jiang, Wenbo and Zhang, Guangyao and Shu, Lianjie},
  journal={Renewable Energy},
  volume={171},
  pages={103--115},
  year={2021},
  publisher={Elsevier},
  doi={10.1016/j.renene.2021.01.143}
}

@article{deng2021deep,
  title={A deep transfer learning method based on stacked autoencoder for cross-domain fault diagnosis},
  author={Deng, Ziwei and Wang, Zhuoyue and Tang, Zhaohui and Huang, Keke and Zhu, Hongqiu},
  journal={Applied Mathematics and Computation},
  volume={408},
  pages={126318},
  year={2021},
  publisher={Elsevier},
  doi={10.1016/j.amc.2021.126318}
}

@article{lei2020applications,
  title={Applications of machine learning to machine fault diagnosis: A review and roadmap},
  author={Lei, Yaguo and Yang, Bin and Jiang, Xinwei and Jia, Feng and Li, Naipeng and Nandi, Asoke K},
  journal={Mechanical systems and signal processing},
  volume={138},
  pages={106587},
  year={2020},
  publisher={Elsevier},
  doi={10.1016/j.ymssp.2019.106587}
}


\end{document}